**Title: Reinforcement Learning Assisted Oxygen Therapy for COVID-19 Patients Under Intensive Care**

Hua Zheng[1], Jiahao Zhu[1], Wei Xie[1], and Judy Zhong[2]

Affiliations

1. Department of Mechanical and Industrial Engineering, Northeastern University, 360 Huntington Avenue, Boston, MA 02115
2. Division of Biostatistics, Department of Population Health, New York University School of Medicine, 180 Madison Avenue, New York, NY 10016

**Corresponding author:**

Wei Xie
Assistant Professor
Mechanical and Industrial Engineering, Northeastern University
360 Huntington Avenue, 334 SN, Boston, MA 02115,
Tel: 6173732740
w.xie@northeastern.edu

and

Judy Zhong
Associate Professor and Director
Division of Biostatistics Department of Population Health
NYU Langone Health
180 Madison Avenue, 4th Floor, Room 452 New York, NY 10016
Tel: 6465013646
judy.zhong@nyulangone.org

**ABSTRACT**

**Background:** Patients with severe Coronavirus disease 19 (COVID-19) typically require supplemental oxygen as an essential treatment. We developed a machine learning algorithm, based on deep Reinforcement Learning (RL), for continuous management of oxygen flow rate for critically ill patients under intensive care, which can identify the optimal personalized oxygen flow rate with strong potentials to reduce mortality rate relative to the current clinical practice.

**Methods:** We modeled the oxygen flow trajectory of COVID-19 patients and their health outcomes as a Markov decision process. Based on individual patient characteristics and health status, an optimal oxygen control policy is learned by using deep deterministic policy gradient (DDPG) and real-time recommends the oxygen flow rate to reduce the mortality rate. We assessed the performance of proposed methods through cross validation by using a retrospective cohort of 1,372 critically ill patients with COVID-19 from New York University Langone Health ambulatory care with electronic health records from April 2020 to January 2021.

**Results:** The mean mortality rate under the RL algorithm is lower than the standard of care by 2.57% (95% CI: 2.08-3.06) reduction (P<0.001) from 7.94% under the standard of care to 5.37 % under our proposed algorithm. The averaged recommended oxygen flow rate is 1.28 L/min (95% CI: 1.14-1.42) lower than the rate delivered to patients. Thus, the RL algorithm could potentially lead to better intensive care treatment that can reduce the mortality rate, while saving the oxygen scarce resources. It can reduce the oxygen shortage issue and improve public health during the COVID-19 pandemic.

**Conclusions:** A personalized reinforcement learning oxygen flow control algorithm for COVID-19 patients under intensive care showed a substantial reduction in 7-day mortality rate as compared to the standard of care. In the overall cross validation cohort independent of the training data, mortality was lowest in patients for whom intensivists' actual flow rate matched the RL decisions.

**Background**

Over the course of the past year, the rapid global spread of severe acute respiratory syndrome coronavirus-2 (SARS-CoV-2), has motivated multidisciplinary investigation efforts to identify effective medical management against coronavirus disease 2019 (COVID-19). Respiratory distress, including mild or moderate respiratory distress, acute respiratory distress syndrome (ARDS) and hypoxia, is a common complication of COVID-19 patients. The therapy of COVID-19 is guided by the knowledge and experience of moderate-to-severe ARDS treatment [1]. Oxygen therapy is recommended as the first-line therapy of COVID-19-induced respiratory and hypoxia by the Centers for Disease Control and Prevention (CDC) and the World Health Organization (WHO). Oxygen therapy consists of different kinds of supplemental oxygen therapies including nasal cannula, simple mask, venturi mask, non-rebreather masks, and high flow oxygen systems. The key factor in different supplemental oxygen methods is the setting of different levels of oxygen flow rates [2]. Thus, the selection of appropriate oxygen flow rate is a crucial decision in COVID-19 treatment. To improve the treatment efficiency, the administration of oxygen therapy should be determined by the severity of COVID-19-induced respiratory failure, incorporating the uncertainties in measurements of patient health status and prediction of individual outcomes to the oxygen decisions. It certainly requires a comprehensive investigation of the optimal and personalized oxygen flow rate. Our research aims to explore effective oxygen therapy for COVID-19 patients based on continuous respiratory support and vital signs monitoring.

A large collection of artificial intelligence (AI) and deep learning (DL) approaches have been proposed to accelerate the drug discovery and the process of diagnosis and treatment of COVID-19 disease [3, 4]. Clinical studies in oxygen therapy and respiratory support have been made in a short period in the treatment of COVID-19 pneumonia [5, 6]. However, respiratory failure remains the leading cause of death (69.5%) for SARS-CoV-2 [7]. Thus, we provide an AI algorithm for the oxygen flow control, based on the deep deterministic policy gradient (DDPG) [8], a widely used reinforcement learning (RL) method for continuous state and action spaces. DDPG uses off-policy data and the Bellman equation to learn the Q-function and then utilizes the resulted Q-function (critic network) to learn a deterministic policy (actor network). To stabilize the training, it considers slow-learning target networks, i.e., actor/critic target networks are updated slowly, hence keeping the estimated targets stable. The optimized policy can recommend personalized optimal oxygen flow rates for COVID-19 patients based on the knowledge of patient health status estimated from patients' electronic health records (EHRs).

Reinforcement learning has been successfully applied in the past to different healthcare problems such as multimorbidity management [9], HIV therapy [10], cancer treatment [11], and anemia treatment in hemodialysis patients [12]. For critical care, given the large amount and granular nature of electronically recorded data, RL is well suited for providing sequential optimal treatment recommendations and improving health outcomes for new ICU patients [13]. Recent studies include treatment strategies for sepsis in intensive care [14] and personalized regime of sedation dosage and ventilator support for patients in Intensive Care Units (ICUs) [15].

Focusing on RL-based oxygen flow rate control (RL-oxygen), we studied its impact on mortality in COVID-19 patients with respiratory failure. The evolution of patients' ICU histories, including treatments, vitals, and health outcomes, was modeled using a Markov decision process (MDP) [14, 16]. At each decision epoch, based on the state (observed patient characteristics, including age, sex, race, smoking status, BMI, and comorbidity diagnoses, 36 daily observed lab test values, and 6 unique vitals), RL selected an oxygen flow rate (ranged from 0 to 60 L/min) and obtained a reward defined based on patient's 7-day survival. Then, following the oxygen flow rate suggested by RL policy, an estimated mortality rate was predicted to compare with the mortality rate in actual practice.

**Methods**
**Study Design and Participants**
Our research team used a retrospective cohort of the New York University Langone Health (NYULH) EHR data on COVID-19 patients to derive and validate the RL algorithm. Eligible patients had positive COVID-19 PCR test and had oxygen therapy in hospital between March 1st 2020 to January 19th 2021. We excluded COVID-19 patients aged below 50 and not been hospitalized as the lacked consistent documentation of vital signs, treatments, and laboratory tests. This study was approved by the NYULH IRB and the data were de-identified to ensure anonymity.

For each patient, we had access to demographic data, including age, sex, race, ethnicity and smoking status, ICU admits and discharge information, in-hospital living status, comorbidities, treatments, and laboratory test data. The comorbidities, including hyperlipidemia, coronary artery disease, heart failure, hypertension, diabetes, asthma or chronic obstructive pulmonary, dementia and stroke, are defined based on the International Classification of Diseases (ICD)-10 diagnosis codes. To reduce the feature dimensionality, we selected 36 laboratory tests based on two criteria: (1) less than 28% missing values; and (2) COVID-19 related tests and vital signs. In specific, we explore the associations between laboratory tests and COVID-19 based on existing literature and clinical findings. For example, recent studies have shown that a reduced estimated glomerular filtration rate (eGFR), low platelet count, low serum calcium level, increased white blood cell count, Neutrophil-to-lymphocyte ratio (NLR), and red blood cell distribution width-coefficient of variation (RDW-CV) are related to high risk of severity and mortality in patients with COVID-19 [17-21]. Additionally, some research suggests well-controlled blood glucose is associated with the lower mortality in COVID-19 patients with Type-2 diabetes [22] and continuous renal potassium level has correlation of hypokalemia, which is common among patients with COVID-19 [23]. Arterial blood gas analysis, including pH, Oxyhemoglobin saturation ($SaO_2$), oxygen saturation ($SpO_2$), partial pressure of oxygen ($PaO_2$) and bicarbonate ($HCO_3$), is commonly used biomarkers measuring the severity of ARDS [24, 25].

In this study, we employed leave-one-hospital-out validation to evaluate the model performance. The whole dataset was divided into 4 batches by the hospital and then we take one batch as validation set and the rest as training set in each simulation.

**RL algorithm Overview**

We model patient health trajectory and the clinical decisions during a course of intensive care over a period of ICU stay by a Markov decision process (MDP) with state, action, and reward. The state of a patient includes the observed patient demographics, vital signs, and laboratory tests at each time $t$. The action refers to the oxygen flow rate. After a sequence of actions, the patient receives a reward if he/she survives in the next 7 days; otherwise, a penalty to death will be given. The cumulative return is defined as the discounted sum of all rewards of each patient received during the ICU stay. The intrinsic design of RL provides a powerful tool to handle sparse and time-delayed reward signals, which makes them well-suited to overcome the heterogeneity of patient responses to actions and the delayed indications of the efficacy of treatments [14]. The details of state, action, and reward are listed as follows.

- State $s_t$: observed patient's characteristics at each time $t$ with information, including demographics, COVID-19 lab tests, and vital signs.
- Action $a_t$: oxygen flow rate ranged from 0 L/min to 60 L/min.
- Reward $r_t$: the reward of an action at time $t$ is measured by its associated ultimate health outcome given the patient's health state. Similar to [14], we used in-hospital mortality as the system-defined penalty and reward.

When a patient survived, a positive reward was released at the end of the patient's trajectory (i.e., a `reward' of +15); a negative reward (i.e., a `penalty of -15) was issued if the patient died. We find that such a reward function can propagate the final health outcome backward to each decision and intervention over the period so that RL can predict long-term effects and dynamically guide the optimal oxygen flow treatment.

- Discount factor $\gamma$: determines how much the RL agents balance rewards in the distant future relative to those in the immediate future. It can take values between 0 and 1 [16]. After considering the ICU stay tends to be short and conducting side experiments, we chose a value of 0.99, which means that we put nearly as much importance on late deaths as opposed to early deaths for each recommended oxygen flow rate.

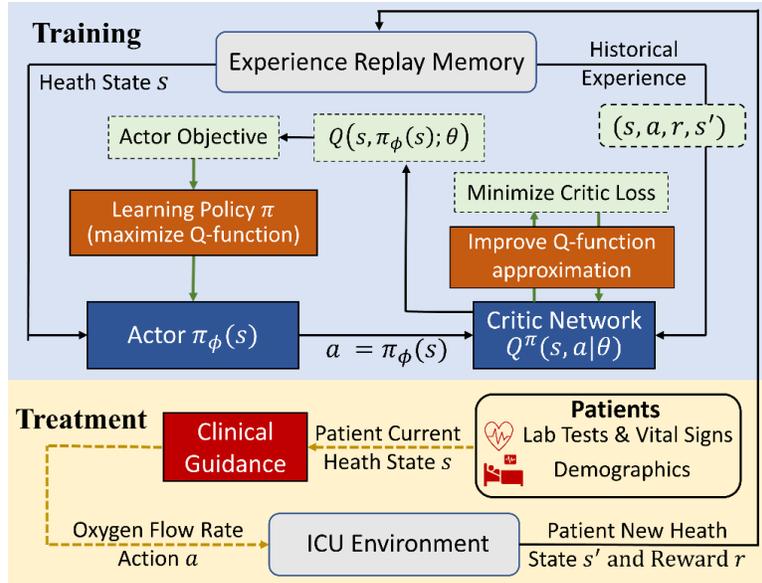

**Fig 1. The diagram of the proposed RL scheme with the actor-critic architecture using electronic health records from New York University Langone Health (NYULH).**

The schematic of the proposed scheme with EHR cohort is shown in Fig. 1. As shown in the bottom part of this diagram, the electronic health data were collected from New York University Langone Health (NYULH) by following the clinical guide. At each time, the oxygen flow rate decision, denoted by $a$, was chosen based on current health state, denoted by $s$, of the patient and then a new heath state $s'$ was observed at the next measurement time. We record the tuple $(s, a, s', r)$ in the experience replay memory with the zero reward $r = 0$. At the end of the treatment, a positive reward was recorded (i.e., a `reward' of +15) if patient survived; and a negative reward (i.e., a `penalty of -15) was issued if the patient died. Then we applied deep deterministic policy gradient (DDPG), as shown at the top part of Fig.

1, to learn the optimal decision policy from the experience replay memory. DDPG, composed of actor and critic networks, takes historical samples $(s, a, s', r)$ from EHR data to concurrently learn a critic network (Q-function approximation) and an actor network (policy). The critic network, denoted by $Q^\pi(s, a|\theta)$, is a nonlinear function that approximates the Q value function

$$Q^\pi(s, a) = \mathrm{E}\left[\sum_{t=0}^{\infty} \gamma^t r_t | s_t = s, a_t = a, \pi\right]$$

of the action $a$ (i.e., oxygen flow rate) given a patient's health state $s$. The actor network representing a policy, denoted by $\pi_\phi(s)$, proposes an action for each given state through the mapping or equation $a = \pi_\phi(s)$. The critic loss, defined by the mean squared TD-error (see Equation (5) in Supplemental Material), is used to improve the approximation of Q-function. Based on the expected Q-value computed by the critic network, we use the policy gradient method to optimize the actor network. In sum, DDPG learns a scoring rule (critic network) which evaluates the performance of a candidate policy, i.e., it returns an oxygen flow rate given a patient's health state, and then uses such a rule to improve the decision making policy (actor network) by optimizing the score. See more details in Reinforcement Learning Algorithms Section of Supplemental Material.

**Model Evaluation**

We evaluated the RL-recommended oxygen therapy by comparing its efficacy with the observed one on the cohort from each validation hospital. At each decision time, the RL algorithm recommends an oxygen flow rate for the patient. If the absolute difference of recommended and the observed oxygen flow rate is less than 10 L/min, we say that RL is "consistent" with the critical care physicians.

When RL is discrepant with the oxygen flow rate used by physicians, the efficacy of the RL-recommended oxygen therapy is not directly observed. The problem then becomes how to assess the health outcomes in the future after taking RL recommendations. For this reason, we predicted the outcome of the RL-recommended treatment using Cox proportional hazards model, a regression model commonly used for investigating the association between the survival probability of patients during a period and predictor variables of interest in medical observational studies [26, 27]. In short, a patient was labeled as "alive" if he/she survived after a treatment within seven days; otherwise, labeled as "deceased". Then we fitted a Cox survival model with demographics, vital signs, and lab tests as predictors and evaluated the effect of decision using the leave-one-hospital-out validation.

To assess the performance of the survival models, we compared predicted and observed outcomes (7-day living status) using 4 metrics: similarity, accuracy, Chi-squared test, and concordance index. Overall, the cosine similarity between predicted and actual survival is greater than 99.9%, and the concordance index is 0.83. Both metrics indicate that the predictive model can effectively estimate unobserved health outcomes. Moreover, the paired Chi-squared test (p-value < 0.0001) shows no significant difference between true and predicted survival.

**Results**

Overall, 1,362 patients in NYULH EHR samples had a PCR-based COVID-19 diagnosis between March 2020 to January 2021. The demographic and clinic characteristics summary of the analysis cohort is shown in Table 1. Overall, patients' mean age is 69.7 and the cohort is comprised of 483 females (35.2%). On average, COVID-19 patients showed BMI of 28.61 kg/m2, pO2 (partial pressure of oxygen) of 104.8 mmHg, SaO2 (Oxygen saturation in arterial blood) of 94.1% and SBP of 123.4 mmHg. Hypertension, hyperlipidemia, diabetes, and coronary artery disease are the top 4 common comorbidities for COVID-19 patients aged above 50, diagnosed in 85.2%, 71.8%, 51.4% and 41.2% patients respectively. The median hospital stay duration was 2.9 days since COVID-19 diagnosis (interquartile range [IQR] 0.52–12.2 days). We trained the RL algorithm using patients from every 3 hospitals and then assessed their performance using the remaining hospital encounters.

The performance of the RL-oxygen is summarized in Table 2. Overall, the RL-oxygen algorithm shows superior performance compared to the clinical practice of oxygen therapy for COVID-19 patients. The overall 7-day estimated mortality under Physician prescribed oxygen was 7.94% (95% CI: 7.41-8.47), while overall estimated mortality under RL-oxygen was 5.37% (95% CI: 4.94-5.80), showing a 2.57% (95% CI: 2.08- 3.06) reduction (P<0.001). In addition, Table 2 depicts the characteristics of oxygen flow rate following the recommendations from both RL-oxygen and physicians. On average, the overall RL-oxygen flow rate was 1.28 L/min (95% CI: 1.14-1.42) lower than the rate delivered to patients.

**Table 1 Demographics and clinical characteristics of NYULH-EHR patients with COVID-19.**

| Demographics and clinic characteristics | Number of Patients (N=1,372) |
|---|---|
| Age (years, Mean (SD)) | 69.72 (10.75) |
| Male (N (%)) | 64.49 (0.47) |
| Race (N(%)) | |
|    African American | 180 (13.12) |
|    Native American | 5 (0.36) |
|    Asian | 120 (8.75) |
|    Caucasian (White) | 730 (53.21) |
|    Multiple Races | 19 (1.39) |
|    Other Races | 266 (19.39) |
|    Race Unknown or Patient Refused | 53 (3.86) |
| Smoking ((N%)) | 1,043 (6.88) |
|    Never | 735 (53.57) |
|    Former | 443 (32.29) |
|    Current | 55 (4.01) |
|    Not asked | 139 (10.13) |
| Body Mass Index (kg/m$^2$, Mean (SD) | 28.61 (6.74) |
| Hyperlipidemia (N(%)) | 978 (71.75) |
| Coronary artery disease (N(%)) | 562 (41.23) |
| Heart failure (N(%)) | 406 (29.79) |
| Hypertension (N(%)) | 1161 (85.18) |
| Diabetes (N(%)) | 701 (51.43) |
| Asthma or chronic obstructive pulmonary (N(%)) | 217 (15.92) |
| Dementia (N(%)) | 133 (9.76) |
| Stroke (N(%)) | 195 (14.31) |

Categorical variables are summarized with frequencies (percentages) unless otherwise indicated. Continuous variables are summarized as the mean (standard deviation) of biomarkers.

The efficacy of the RL prescriptive algorithm was consistently observed across age, gender, BMI, and comorbidity subgroups (Table 2). Demographically speaking, compared to the observed efficacies in patients of age 75 and younger, COVID-19 patients of age older than 75 observed higher efficacies from RL-oxygen recommended therapy than physician's recommendations. For example, 7-day estimated mortality rate under RL-oxygen for patients of age older than 80 was 5.87% (95% CI: 4.67-7.07) lower than under physician's therapy. In contrast, the 7-day estimated mortality rate under RL-oxygen was 0.55% (95% CI: 0.39-0.71) lower than that under physicians' therapy for patients aged between 50 and 65. Table 2 also shows that the RL-oxygen tends to be more effective for patients with comorbidities. Especially for COVID-19 patients with Asthma or chronic obstructive pulmonary, Dementia and Stroke, RL-oxygen reduced the 7-day mortality by 5.69%, 5.11% and 3.8% respectively on average.

**Table 2 Subgroup comparison of 7-day estimated mortality obtained using RL-oxygen algorithm and critical care physician decision guidance.**

| Subgroups | Estimated Mortality (%) | | Average Oxygen (L/min) | |
|---|---|---|---|---|
| | RL-oxygen | Physician | RL-oxygen | Physician |
| Overall | 5.37 (0.22) | 7.94 (0.27) | 19.24 (0.07)* | 20.52 (0.07) |
| Male | 6.13(0.12)* | 8.53(0.14) | 21.20(0.09)* | 22.66(0.09) |
| Female | 2.18(0.11)* | 2.99(0.12) | 6.33(0.07) | 6.41(0.07) |
| Age | | | | |
| 50 to 65 | 1.19(0.08)* | 1.74(0.09) | 25.54(0.12)* | 22.27(0.12) |
| 65 to 75 | 4.13(0.14)* | 5.43(0.16) | 19.63(0.12)* | 22.73(0.12) |
| 75 to 80 | 14.76(0.3)* | 20.39(0.34) | 19.79(0.14)* | 21.45(0.16) |
| ≥80 | 15.86(0.57)* | 21.73(0.65) | 14.28(0.18)* | 18.96(0.26) |
| Body Mass Index (kg/m$^2$) | | | | |
| <25 | 7.74(0.18)* | 11.10(0.21) | 19.27(0.11)* | 20.58(0.12) |
| 25 to 30 | 7.38(0.19)* | 9.21(0.21) | 23.39(0.13)* | 24.5(0.14) |
| 30-35 | 2.72(0.15)* | 5.30(0.21) | 22.91(0.16)* | 21.42(0.17) |
| ≥35 | 5.35(0.28)* | 5.44(0.28) | 19.53(0.18)* | 22.78(0.21) |
| Hyperlipidemia | 7.43(0.13)* | 9.47(0.14) | 20.11(0.09)* | 20.94(0.09) |
| Coronary artery disease | 8.55(0.18)* | 11.39(0.21) | 18.13(0.12)* | 20.04(0.11) |
| Heart failure | 11.25(0.23)* | 12.59(0.25) | 18.35(0.11) | 18.22(0.13) |
| Hypertension | 6.96(0.11)* | 8.79(0.13) | 21.2(0.08) | 21.25(0.08) |
| Diabetes | 7.73(0.15)* | 8.25(0.15) | 25.22(0.11)* | 20.31(0.1) |
| Asthma or chronic obstructive pulmonary | 11.98(0.32)* | 17.67(0.38) | 15.57(0.15)* | 19.68(0.18) |
| Dementia | 10.71(0.46)* | 15.82(0.56) | 15.57(0.23)* | 14.19(0.23) |
| Stroke | 9.15(0.31)* | 12.95(0.37) | 21.78(0.15)* | 15.94(0.19) |

Categorical variables are summarized with frequencies (percentages) unless otherwise indicated. Continuous variables are summarized as the mean (standard error) of biomarkers.

*Variables indicate RL-oxygen is significantly different from physicians (p-value<0.001).

We further studied 7-days mortality when the actually administered oxygen flow rate differed from the oxygen flow rate suggested by the RL-oxygen in Fig. 2. It shows how the observed mortality changes with the flow rate difference between RL-oxygen and physicians. This phenomenon suggests that increasing differences between the RL-oxygen and the observed delivering oxygen were associated with increasing observed mortality rates in a rate-dependent fashion. When the difference is minimum, we obtain the lowest 7-day mortality rate of 1.7%. Another observation from Fig. 2(A) is that the mortality rate increases when the RL-oxygen flow rate is lower or higher than the one from physicians. It suggests that both the oxygen deficit (lower oxygen flow rate than RL-oxygen recommendation) and the oxygen excess are sup-optimal for patients' outcomes. We observed a trend that RL-oxygen was in general lower than what was prescribed by the physicians and might result in better outcomes under a lower flow rate. It suggests that oxygen flow rates prescribed by doctors tend to be excessively high for some patients.

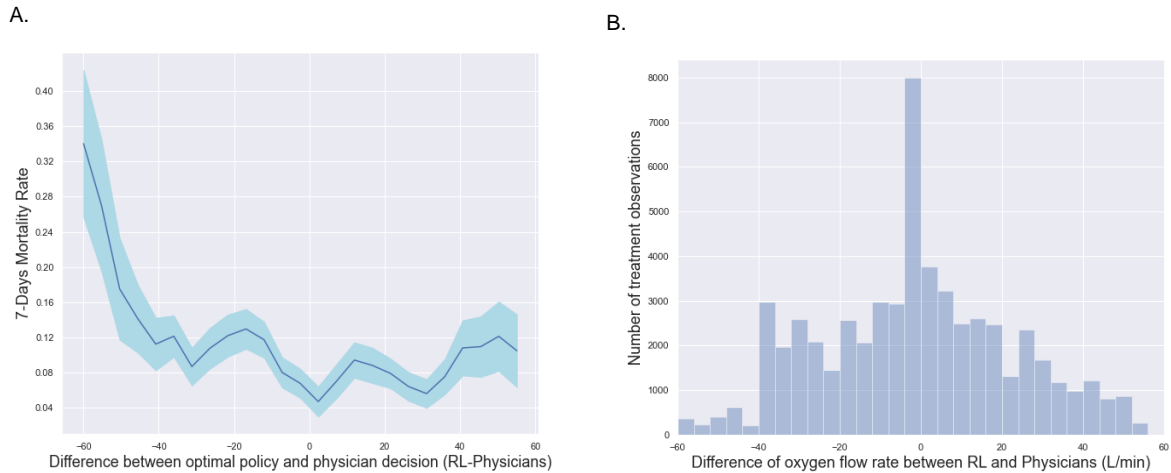

**Fig 2. (A) Comparison of the estimated 7-days mortality rates (y-axis) varying with the difference between the oxygen flow rate recommended by the RL optimal policy and that administered by doctors (x-axis) averaged over all time points per patient. The shaded area represents the 95% confidence interval. The smallest oxygen difference is mainly associated with the lowest 7-days mortality rates. The further away the dose received was from the suggested oxygen flow rate, the worse the outcome. (B) The histogram of oxygen flow rate difference between RL-oxygen and physicians (labels on the vertical axis).**

Last, we observed that the RL-oxygen and physicians recommended consistent flow rates in most times; see Fig. 2(B). The overall distribution of oxygen flow rates recommended by RL-oxygen and physicians are presented in Fig. 3. It depicts how many measurement times each oxygen flow rate was recommended by RL-oxygen and physicians. In twenty-nine percent of the time, the patients received an oxygen flow close to the suggested rate within 5 L/min while forty-four percent of the time, the difference between the administered and suggested oxygen flow rates are within 10L/min. Since the high-flow nasal oxygen (HFNO) therapy often increases flow rate in increments of 10 L/min up to 60 L/min [28], it suggests that RL-oxygen is consistent with physicians about 40-50% of the time.

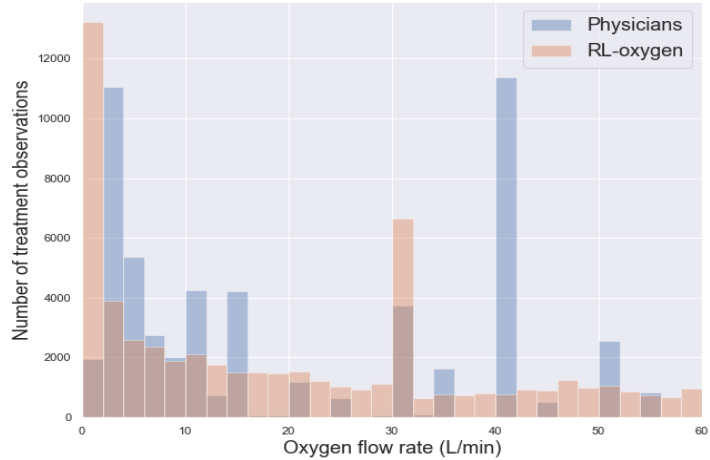

**Fig 3. Oxygen delivery by RL versus critical care physicians.** Histogram of oxygen flow rate delivered to COVID-19 patients; blue bar indicates physician and orange bar indicates RL-oxygen.

**Discussion**

We used a RL approach to learn an optimal policy to continuously control the oxygen device for critically ill patients with COVID-19 who require oxygen therapy. As most people who become seriously unwell with COVID-19 have an acute respiratory illness [29, 30], our algorithm has strong potential to improve individual health outcomes and reduce the COVID-19 mortality rate caused by respiratory failure. We designed the reward as the ultimate health outcome which is used to assess the performance of oxygen flow decisions along the treatment trajectory. As such, the reinforcement learning approach took uncertain outcomes and long-term treatment effects into consideration and made it smarter in understanding the long impact of an early decision on the final outcomes.

Our analysis suggests the current practice remains some potential to be improved as actual oxygen flow rate administered by intensivists showed more than fifty percent discrepancy with RL-oxygen recommendations. Importantly, we observe that RL-oxygen tends to prescribe lower oxygen flow rate than physician's prescribed rates but leads to better outcomes. This finding is especially important in the context of the ongoing and persistent medical oxygen shortages in some developing regions. As COVID-19 patient-care protocols have evolved, medical-grade oxygen is still considered as an essential resource to treatments for critically ill patients. In regions such as Africa, Middle East, and Asia, the surge in demand for medical oxygen to treat COVID-19 exacerbates preexisting gaps in medical-oxygen supplies, leading to substantial supply shortages.

Our analysis also identified some clinical patterns that RL-oxygen particularly works well. For example, patients with high risk (i.e., of age older than 75) observed higher efficacies than patients aged between 50 and 75 by using relatively lower averaged oxygen flow rate than actually administered. RL-oxygen also recommends a higher averaged oxygen flow rate may improve the health outcomes for patients aged from 50 to 65. Moreover, we also notice significant therapeutic discrepancies in patients with stroke and diabetes comorbidities. In both cases, RL-oxygen recommended higher averaged oxygen flow rate than doctors while showing a significant reduction in estimated mortality. In fact, these findings agree with recent studies which reported that "stroke survivors who underwent COVID-19 developed more acute respiratory distress syndrome and received more noninvasive mechanical ventilation" [31] and "diabetic patients required more oxygen therapy (60% versus 26.9%)" [32].

Although our evaluation methodology controls for several confounding factors and shows high validation accuracy, sample scarcity and a large proportion of missing value may increase estimation uncertainty and affect the treatment recommendations. Larger training data are necessary to cover more of the state space and improve the policy optimization. Moreover, the COVID-19 cohort from NYULH may not be representative of the U.S. COVID-19 population or the oxygen clinical practices in other countries. To ultimately validate the efficacy of the RL algorithm, randomized clinical trials with patients randomly assigned to RL and clinician oxygen therapy would be needed.

**Conclusion**

Through analyzing the EHR data from multiple ambulatory care centers, we demonstrated the feasibility of using reinforcement learning based oxygen therapy to improve the intensive care for COVID-19 patients. The RL-oxygen showed medium concordance (44%) with the current practice of critical care physicians. For all COVID-19 patients requiring oxygen therapy, RL recommendations significantly reduce the mortality rate compared to the current practice. The algorithm has the potential to be integrated into the clinical decision support system and assist physicians to provide timely personalized recommendations of oxygen flow rate for COVID-19 patients in ICU.

**References**

1. Tzotzos SJ, Fischer B, Fischer H, Zeitlinger M: **Incidence of ARDS and outcomes in hospitalized patients with COVID-19: a global literature survey**. *Critical Care* 2020, **24**(1):1-4.
2. Whittle JS, Pavlov I, Sacchetti AD, Atwood C, Rosenberg MS: **Respiratory support for adult patients with COVID-19**. *Journal of the American College of Emergency Physicians Open* 2020, **1**(2):95-101.
3. Jamshidi MB, Lalbakhsh A, Talla J, Peroutka Z, Hadjilooei F, Lalbakhsh P, Jamshidi M, Spada L, Mirmozafari M, Dehghani M *et al*: **Artificial Intelligence and COVID-19: Deep Learning Approaches for Diagnosis and Treatment**. *IEEE Access* 2020, **8**:109581-109595.
4. Jamshidi MB, Lalbakhsh A, Talla J, Peroutka Z, Roshani S, Matousek V, Roshani S, Mirmozafari M, Malek Z, La Spada L: **Deep Learning Techniques and COVID-19 Drug Discovery: Fundamentals, State-of-the-Art and Future Directions**. *Emerging Technologies During the Era of COVID-19 Pandemic* 2021, **348**:9.
5. Marini JJ, Gattinoni L: **Management of COVID-19 respiratory distress**. *Jama* 2020, **323**(22):2329-2330.
6. Attaway AH, Scheraga RG, Bhimraj A, Biehl M, Hatipoğlu U: **Severe covid-19 pneumonia: pathogenesis and clinical management**. *BMJ* 2021, **372**:n436.
7. Zhang B, Zhou X, Qiu Y, Song Y, Feng F, Feng J, Song Q, Jia Q, Wang J: **Clinical characteristics of 82 cases of death from COVID-19**. *PloS one* 2020, **15**(7):e0235458.
8. Lillicrap TP, Hunt JJ, Pritzel A, Heess N, Erez T, Tassa Y, Silver D, Wierstra D: **Continuous control with deep reinforcement learning**. *arXiv preprint arXiv:150902971* 2015.
9. Zheng H, Ryzhov IO, Xie W, Zhong J: **Personalized Multimorbidity Management for Patients with Type 2 Diabetes Using Reinforcement Learning of Electronic Health Records**. *Drugs* 2021:1-12.
10. Ernst D, Stan G, Goncalves J, Wehenkel L: **Clinical data based optimal STI strategies for HIV: a reinforcement learning approach**. In: *Proceedings of the 45th IEEE Conference on Decision and Control: 13-15 Dec. 2006 2006*; 2006: 667-672.
11. Zhao Y, Zeng D, Socinski MA, Kosorok MR: **Reinforcement Learning Strategies for Clinical Trials in Nonsmall Cell Lung Cancer**. *Biometrics* 2011, **67**(4):1422-1433.
12. Escandell-Montero P, Chermisi M, Martínez-Martínez JM, Gómez-Sanchis J, Barbieri C, Soria-Olivas E, Mari F, Vila-Francés J, Stopper A, Gatti E *et al*: **Optimization of anemia treatment in hemodialysis patients via reinforcement learning**. *Artificial Intelligence in Medicine* 2014, **62**(1):47-60.
13. Liu S, See KC, Ngiam KY, Celi LA, Sun X, Feng M: **Reinforcement learning for clinical decision support in critical care: comprehensive review**. *Journal of medical Internet research* 2020, **22**(7):e18477.
14. Komorowski M, Celi LA, Badawi O, Gordon AC, Faisal AA: **The Artificial Intelligence Clinician learns optimal treatment strategies for sepsis in intensive care**. *Nat Med* 2018, **24**(11):1716-1720.
15. Prasad N, Cheng L-F, Chivers C, Draugelis M, Engelhardt BE: **A Reinforcement Learning Approach to Weaning of Mechanical Ventilation in Intensive Care Units**. *arXiv* 2017.
16. Sutton RS, Barto AG: **Reinforcement Learning: An Introduction**, Second edn: The MIT Press; 2018.
17. Lippi G, Plebani M, Henry BM: **Thrombocytopenia is associated with severe coronavirus disease 2019 (COVID-19) infections: a meta-analysis**. *Clinica chimica acta* 2020, **506**:145-148.
18. Moradi EV, Teimouri A, Rezaee R, Morovatdar N, Foroughian M, Layegh P, Kakhki BR, Koupaei SRA, Ghorani V: **Increased age, neutrophil-to-lymphocyte ratio (NLR) and white blood cells count are associated with higher COVID-19 mortality**. *The American Journal of Emergency Medicine* 2021, **40**:11-14.
19. Zhou X, Chen D, Wang L, Zhao Y, Wei L, Chen Z, Yang B: **Low serum calcium: a new, important indicator of COVID-19 patients from mild/moderate to severe/critical**. *Bioscience reports* 2020, **40**(12).
20. Wang C, Deng R, Gou L, Fu Z, Zhang X, Shao F, Wang G, Fu W, Xiao J, Ding X: **Preliminary study to identify severe from moderate cases of COVID-19 using combined hematology parameters**. *Ann Transl Med* 2020, **8**(9).
21. Cheng Y, Luo R, Wang K, Zhang M, Wang Z, Dong L, Li J, Yao Y, Ge S, Xu G: **Kidney impairment is associated with in-hospital death of COVID-19 patients**. *MedRxiv* 2020.
22. Zhu L, She Z-G, Cheng X, Qin J-J, Zhang X-J, Cai J, Lei F, Wang H, Xie J, Wang W: **Association of blood glucose control and outcomes in patients with COVID-19 and pre-existing type 2 diabetes**. *Cell metabolism* 2020, **31**(6):1068-1077. e1063.

23. Chen D, Li X, Song Q, Hu C, Su F, Dai J, Ye Y, Huang J, Zhang X: **Assessment of hypokalemia and clinical characteristics in patients with coronavirus disease 2019 in Wenzhou, China**. *JAMA network open* 2020, **3**(6):e2011122-e2011122.
24. Rice TW, Wheeler AP, Bernard GR, Hayden DL, Schoenfeld DA, Ware LB, Network A, Health NIo: **Comparison of the SpO2/FIO2 ratio and the PaO2/FIO2 ratio in patients with acute lung injury or ARDS**. *Chest* 2007, **132**(2):410-417.
25. Chen W, Janz DR, Shaver CM, Bernard GR, Bastarache JA, Ware LB: **Clinical characteristics and outcomes are similar in ARDS diagnosed by oxygen saturation/Fio2 ratio compared with Pao2/Fio2 ratio**. *Chest* 2015, **148**(6):1477-1483.
26. Cummings MJ, Baldwin MR, Abrams D, Jacobson SD, Meyer BJ, Balough EM, Aaron JG, Claassen J, Rabbani LE, Hastie J: **Epidemiology, clinical course, and outcomes of critically ill adults with COVID-19 in New York City: a prospective cohort study**. *The Lancet* 2020, **395**(10239):1763-1770.
27. Bradburn MJ, Clark TG, Love SB, Altman DG: **Survival analysis part II: multivariate data analysis–an introduction to concepts and methods**. *British journal of cancer* 2003, **89**(3):431-436.
28. Ho C-H, Chen C-L, Yu C-C, Yang Y-H, Chen C-Y: **High-flow nasal cannula ventilation therapy for obstructive sleep apnea in ischemic stroke patients requiring nasogastric tube feeding: a preliminary study**. *Scientific Reports* 2020, **10**(1):1-8.
29. Wu Z, McGoogan JM: **Characteristics of and Important Lessons From the Coronavirus Disease 2019 (COVID-19) Outbreak in China: Summary of a Report of 72 314 Cases From the Chinese Center for Disease Control and Prevention**. *JAMA* 2020, **323**(13):1239-1242.
30. Nicholson TW, Talbot NP, Nickol A, Chadwick AJ, Lawton O: **Respiratory failure and non-invasive respiratory support during the covid-19 pandemic: an update for re-deployed hospital doctors and primary care physicians**. *BMJ* 2020, **369**:m2446.
31. Qin C, Zhou L, Hu Z, Yang S, Zhang S, Chen M, Yu H, Tian DS, Wang W: **Clinical Characteristics and Outcomes of COVID-19 Patients With a History of Stroke in Wuhan, China**. *Stroke* 2020, **51**(7):2219-2223.
32. Elamari S, Motaib I, Zbiri S, Elaidaoui K, Chadli A, Elkettani C: **Characteristics and outcomes of diabetic patients infected by the SARS-CoV-2**. *Pan Afr Med J* 2020, **37**:32.

**Title: Reinforcement Learning Assisted Oxygen Therapy for COVID-19 Patients Under Intensive Care**

## Supplemental Material

**Cox Proportional Hazards Model**

Cox proportional hazards model [1] is a regression model commonly applied for investigating the association between the risk factors and survival time of patients. Its primary output is the mortality rate of patients. In this study, researchers used multivariable Cox PH model to predict the mortality rate. The input variables in the Cox PH model include: (1) decision, i.e., the oxygen flow rate of the oxygen therapy; and (2) the risk factors associated to COVID-19, such as age, hypertension, and diabetes. Besides, we set up a 7-day time window to estimate the mortality rate and evaluate the efficacy of a given oxygen therapy. Specifically, we formalize the variables in Cox PH model as follows.

- $t$: denotes the duration since the patient admission time;
- $q(t)$: denotes the survival rate of patients at time $t$;
- $s$: denotes the health state as predictor variables related to the survival rate;
- $\beta$: denotes the coefficient of the corresponding variables.

The objective of Cox PH model used to predict the survival rate is given by

$$q(t \mid s) = \exp\left(-\int_0^t \lambda(z|s)dz\right), \tag{1}$$

where the hazard at time $t$ for an individual with health state $s$ is assumed to be

$$\lambda(t \mid s) = \lambda_0(t)\exp(s^\top \beta).$$

In this model, $\lambda_0(t)$ is a baseline hazard function, and $\exp(s^\top \beta)$ is the relative risk, a proportionate increase or reduce in risk, associated with the set of characteristics $s$. Note that the increase or reduce in risk is the same at all duration t. Given health state $s$ of a patient, we predict the 7-day mortality rate by using $1 - q(7 \mid s)$.

**Feature Selection**

Feature selection is critical for the establishment of Cox PH model. Unrelated risk factors and the high multilinearity between predictor variables will cause low concordance and impact on the prediction. In addition to the selected 36 laboratory tests (see **Study Design and Participants**), we also included 25 additional demographic predictor variables, 2 vital signs (temperature and systolic blood pressure), and oxygen flow rate. Since there were high correlations between the selected features, we conduct feature selection based on Pearson correlation to preclude multilinear features. Basically, we found the high linear correlation (>0.7) existing in each group of features, including: (1) red blood cell distribution width-coefficient of variation (RDW-CV) and red cell volume distribution width-standard deviation (RDW-SD); (2) eGFR and creatinine; (3) red blood cell count, hemoglobin, and hematocrit; (4) neutrophils and lymphocytes; and (5) SpO2, oxyhemoglobin and methemoglobin. We selected the first predictor in each group and removed the rest: RDW-CV, eGFR, red blood cell count, neutrophils and SpO2.

For the rest feature selection, we used the elastic net regularization [2] with grid search [3] to select the features. The procedure is shown as follows.

1. We create a grid of possible values for regularizers in cross-product of L1 and L2 penalty values ranging in [0.01, 0.02, 0.04, 0.06, 0.08]. It results in 25 different combinations in total, i.e., (0.01,0.01), (0.01,0.02) ... , (0.08,0.06), (0.08,0.08).
2. For each combination of L1 and L2 penalty values, we fitted a Cox model with elastic net regularization and recorded the performance measured by the concordance score.
3. Finally, we chose the best L1 and L2 regularizers with the best performance.

In the study, the selected coefficients of L1 and L2 regularizers are 0.04 and 0.02 respectively.

**Training Process**

We apply leave-one-hospital-out cross validation to evaluate the models and predict the 7-day survival probability to assess the performance of RL-oxygen models. To train the general model (including data from different hospitals), we randomly select 80% of the cohort as training set and the rest 20% as the test set. The coefficient of predictor variables of general Cox PH model shows in table S.1.

**Table S.1 Selected features for Cox proportional-hazards model.**

| Feature name | Coefficient | SE | 95% CI | p-value |
|---|---|---|---|---|
| Age, years | 0.02 | 0.00 | 0.02 | <0.001 |
| Anion gap, mEq/L | 0.03 | 0.02 | [ 0.02, 0.03] | <0.001 |
| Blood urea nitrogen, mg/dL | 0.00 | 0.00 | 0.00 | <0.001 |
| Serum calcium, mg/dL | -0.19 | 0.01 | [-0.22, -0.16] | <0.001 |
| PaCO2, mm Hg | -0.01 | 0.00 | [-0.02, -0.01] | <0.001 |
| Eosinophils, cells/μL | -0.04 | 0.01 | [-0.05, -0.02] | <0.001 |
| HCO3, mEq/L | -0.01 | 0.00 | [-0.02, -0.01] | <0.001 |
| Mean platelet volume, fL | 0.05 | 0.01 | [0.03, 0.07] | <0.001 |
| Nucleated red blood count, /100 WBC | 0.08 | 0.02 | [0.04, 0.12] | <0.001 |
| PH | -1.86 | 0.11 | [-2.08, -1.64] | <0.001 |
| Inorganic phosphorus, mg/dL | 0.03 | 0.01 | [0.02, 0.05] | <0.001 |
| PaO2, mmHg | 0.00 | 0.00 | 0.00 | <0.001 |
| Potassium, mEq/L | 0.15 | 0.02 | [0.11, 0.18] | <0.001 |
| RDW-CV, % | 0.06 | 0.00 | [0.05, 0.06] | <0.001 |
| White blood cell count | 0.01 | 0.00 | 0.01 | <0.001 |
| Oxygen flow rate, L/min | 0.01 | 0.00 | 0.01 | <0.001 |

The confidence interval is replaced by the coefficient estimates if the SE is smaller than 0.01

## Reinforcement Learning Algorithm

A Markov decision process (MDP) was used to model the decision-making process and approximate individual patient health trajectories. We formalize the MDP by the tuple $(S, A, P, r, \gamma)$.

- $S$: denotes a finite set of states, typically including patients' demographic information, ICU admit and discharge information, comorbidities, treatment and laboratory tests.
- $A$: denotes action space that includes oxygen flow rate over time. In this problem, we consider continuous oxygen flow rate $a_t \in A$.
- $P(s'|a, s)$: represents the state transition probability model that takes action $a$ in state $s$ at time $t$ and will lead to the transition to state $s'$ at next time $t + 1$ (i.e., the patient's health state changes to $s'$ at $t + 1$ after taking oxygen therapy with flow rate $a$ at time $t$), which describes the dynamics of the treatment and health interactions.
- $r$: represents the immediate reward received for transitioning to state $s'$. Transitions to desirable states yield a positive reward, and reaching undesirable states generates a penalty.
- $\gamma$: denotes the discount factor, which makes immediate rewards more valuable than long-term rewards and determines the temporal impact of the current action. The greater $\gamma$ indicates longer impact of current therapy action.

The process is observed at discrete time steps. In each time $t$, the agent observes the current state $s_t \in S$. Then, we choose an action $a_t \in A$ (i.e., oxygen flow rate), the patient health condition moves to a new state $s_{t+1}$, and we get a reward signal $r_{t+1}$ associated with the one-step transition $(s_t, a_t, s_{t+1})$. The oxygen flow rate decision making strategy is called the *policy*, denoted by a mapping $\pi$ from state space $S$ to action space $A$, i.e., $a_t = \pi(s_t)$. The performance of a policy is measured using the value function

$$V^\pi(s) = E[\sum_{t=0}^{\infty} \gamma^t r_t | s_t = s, \pi] \quad (2)$$

which is defined as the expected cumulative discounted reward starting with state $s$, given that policy $\pi$ is used to make decisions. Then, the goal of a reinforcement learning agent is to learn the optimal policy $\pi^*$ which maximizes the expected cumulative discounted reward, that is, $V^\pi(s)$.

The reward $r_t(s_t, s_{t+1})$ at each time $t$ is defined as follows,
- If patient stays alive, $r_t = 0$;
- If patient is discharged, $r_t = 15$;
- If patient died, $r_t = -15$;

**Learning the Optimal Policy**

We utilize the Deep Deterministic Policy Gradient (DDPG) to concurrently learn the Q-function and optimal policy. In each iteration, we use off-policy data and the Bellman equation to learn the Q-function, and then the estimated Q-function is used to learn the optimal policy.

This approach is closely connected to *Q-learning*. In reinforcement learning, many algorithms focus on estimating the so-called "Q-function", denoted by $Q^\pi(s, a)$, of a policy $\pi$. The Q-function measures the expected return or discounted sum of rewards obtained by following the policy $\pi$ and acting $a = \pi(s)$. Thus, the Q-function represents the expected value of state-action pairs, and it can be connected to the value function through the equation

$$V^\pi(s) = \max_a Q^\pi(s, a). \quad (3)$$

DDPG interleaves the learning process for a good approximator to $Q^{\pi^*}(s, a)$ with the learning process for an approximator to the optimal policy $\pi^\star(s)$. The *optimal* Q-function is then defined as the maximum return that can be obtained starting from state $s$, acting $a$, and following the optimal policy $\pi^*$ thereafter. It is known to obey the following Bellman optimality equation:

$$Q^{\pi^*}(s, a) = E_{s'}[r(s, a) + \gamma \max_{a'} Q^{\pi^*}(s', a')] \quad (4)$$

where the next state $s'$ is sampled from the state transition distribution, denoted by $P(\cdot | s, a)$. For continuous action space, the function $Q^\pi(s, a)$ is presumed to be differentiable with respect to the action argument.

We use a nonlinear function, such as a neural network with parameters $\theta$, to approximate the state-action value function, i.e., $Q^\pi(s, a) \approx Q^\pi(s, a; \theta)$. Such a neural network is called a Q-network [29]. Let $a(s) = \pi_\phi(s)$ denote the deterministic policy function parameterized by $\phi$. The Q-function is trained by minimizing the approximation difference (**critic loss** function) between the left- and right-hand side in Eq. (4), i.e.,

$$L(s,a) = \frac{1}{2} E_{s' \sim p(\cdot|s,a)} \left[ \left( Q^\pi(s,a;\theta) - r(s,a) - \gamma \max_\pi Q^\pi\left(s', \pi_{\tilde{\phi}}(s'); \tilde{\theta}\right) \right)^2 \right], \quad (5)$$

or equivalently,

$$L(s,a) = E_{s' \sim p(\cdot|s,a)} [\ell_\theta(s, a, \pi_{\tilde{\phi}}(s'))], \quad (6)$$

with

$$\ell_\theta(s,a,s') = \frac{1}{2} \left( Q^\pi(s,a;\theta) - r(s,a) - \gamma \max_\pi Q^\pi\left(s', \pi_{\tilde{\phi}}(s'); \tilde{\theta}\right) \right)^2$$

where $\tilde{\theta}$ denotes the target Q-function parameters and $\tilde{\phi}$ denotes the target policy function parameters. Both parameter values $\tilde{\theta}$ and $\tilde{\phi}$ are obtained from the last iteration. We call

$$target(s,a,s') = r(s,a) + \gamma Q^\pi\left(s', \pi_{\tilde{\phi}}(s'); \tilde{\theta}\right)$$

as the target value and $Q^\pi(s,a;\theta) - target(s,a,s')$ as **TD error**. Ideally, we want the error to decrease, meaning that our current policy's outputs are becoming closer to the true Q values. Then, by differentiating the loss function with respect to the parameters $\theta$, we have the gradient,

$$\nabla_\theta \ell_\theta(s,a,s') = \left( Q^\pi(s,a;\theta) - \text{target}(s,a,s') \right) \nabla_\theta Q^\pi(s,a;\theta). \quad (7)$$

The **policy learning** step in DDPG will obtain a deterministic policy $\pi_\phi(s)$ which gives the action maximizes $Q(s,a;\theta)$. Because the action space is continuous, we assume the Q-function is differentiable with respect to action parameters. Considering the following discounted expected reward (**actor objective** function)

$$\max_\phi J(\theta) = E_{s \sim \mathcal{D}} [Q(s, \pi_\phi(s); \theta)] \quad (8)$$

we can perform the gradient ascent with respect to policy parameters (see [4] for more details),

$$\nabla_\phi J(\phi) = E[\nabla \pi_\phi(s) \nabla_a Q^\pi(s,a;\theta)]$$

where the expectation is estimated by using the training set, denoted by $D$, of tuple $(s,a,s',r)$ from the EHR data. Then, we update the parameters of the Q-function and the policy function by using the noisy gradient estimates in Eq. (7) and (8) and obtain new parameters $\theta$ and $\phi$. At the end of each iteration, we update the target network, i.e., Q function $Q^\pi\left(s', \pi_{\tilde{\phi}}(s'); \tilde{\theta}\right)$ in $target(s,a,s')$ and target policy by

$$\tilde{\theta} \leftarrow \rho \tilde{\theta} + (1-\rho)\theta$$
$$\tilde{\phi} \leftarrow \rho \tilde{\phi} + (1-\rho)\phi$$

where $\rho$ is a hyperparameter between 0 and 1.

The Q-network model, a.k.a. critic network in our paper, uses a multi-layer feed-forward architecture which evaluates each state-action pair $(s,a)$. Specifically, the model architecture contains a state input layer followed by a dense layer with 32 neurons and an action input layer; they are concatenated and then followed by a 16-dimensional dense layer; the output layer is 1 dimensional with a linear activation function. The policy model, a.k.a. actor network, uses a two-later neural network with the state input followed by 32-dimensional intermediate layer and 1-dimensional action output layer. We also use batch normalization [5] after each dense layer to standardize the unit of low dimensional features. It is particularly useful in healthcare data as most biomarkers and vital signs have different physical unit and characteristics by nature and even statistics of the same type may vary a lot across multiple patients. Batch normalization can fix this issue by normalizing every dimension across samples in one minibatch.

We used the early stopping [6, 7] to prevent overfitting. There are two metrics used as early stopping criteria: mean squared TD error and consistency of recommendations between physician and RL. First, since the objective of DDPG is to minimize the mean squared TD error $(7)$, it is natural to use $(7)$ as a metric. Second, as we did not want RL-oxygen to be too much different from the standard of care, we used the consistency of recommendations as another metric, which is defined by the mean square deviations between RL's and physicians' recommended oxygen flow rates. In the study, we noticed that this second metric tends to converge later than the TD error. Thus, during training, we monitored both metrics and set the early stopping criterion to be that "mean squared deviation is not improved in last 500 iterations".

Our training scheme is as follows:
1. Split the dataset into 4 groups (one hospital per fold)
2. For each unique group:
    1) Take the group as a hold out or test data set;
    2) Take the remaining groups as a training data set;
    3) Fit a model on the training set and evaluate it on the test set;
    4) Retain the evaluation score;
    5) Repeat this process until every group serves as the test set.
3. Then take the average of the recorded scores as the performance metric for the model.

In reinforcement learning, learning an optimal policy from observational data is referred as to offline RL [13]. This approach uses a set of one-step transition tuples: $D = \{(s_i, a_i, r_i, s'_i): i = 1, ..., |D|\}$ to estimate the Q-function $Q^\pi(s, a'; \theta)$ and the oxygen flow policy $\pi(s)$. The learning algorithm follows [23] with 64 batch size and 0.002 learning rates for both critic and actor network.

**Missing Data Imputation**

Our dataset contains a set of historically observed health states, but not every possible health state, and the time series data such as lab tests, vital signs, and oxygen flow rate are sampled unevenly. To learn an optimal policy, RL requires a way to estimate values in any state, including those not in the original data. Thus, we imputed data for such missing states based on the information from nearby measurements using a linear interpolation method.

**References**

1. Cox DR: **Regression models and life-tables**. *Journal of the Royal Statistical Society: Series B (Methodological)* 1972, **34**(2):187-202.

2. Zou H, Hastie T: **Regularization and variable selection via the elastic net**. *Journal of the royal statistical society: series B (statistical methodology)* 2005, **67**(2):301-320.
3. LaValle SM, Branicky MS, Lindemann SR: **On the relationship between classical grid search and probabilistic roadmaps**. *The International Journal of Robotics Research* 2004, **23**(7-8):673-692.
4. Lillicrap TP, Hunt JJ, Pritzel A, Heess N, Erez T, Tassa Y, Silver D, Wierstra D: **Continuous control with deep reinforcement learning**. *arXiv preprint arXiv:150902971* 2015.
5. Ioffe S, Szegedy C: **Batch normalization: Accelerating deep network training by reducing internal covariate shift**. In: *International conference on machine learning: 2015*: PMLR; 2015: 448-456.
6. Caruana R, Lawrence S, Giles CL: **Overfitting in neural nets: Backpropagation, conjugate gradient, and early stopping**. In: *Advances in neural information processing systems: 2001*; 2001: 402-408.
7. Yao Y, Rosasco L, Caponnetto A: **On early stopping in gradient descent learning**. *Constructive Approximation* 2007, **26**(2):289-315.